\documentclass[letterpaper, 10 pt, conference]{ieeeconf}  

\IEEEoverridecommandlockouts                              

\usepackage{amsmath} 
\usepackage{amssymb,bm}  

\usepackage{graphics} 
\usepackage{mathptmx} 
\usepackage{times} 

\usepackage{mathptmx}
\usepackage{mathtools}
\usepackage{url}
\usepackage{tikz}
\usepackage[linesnumbered,ruled]{algorithm2e}
\usepackage[tight]{subfigure}
\usepackage{multirow}
\usepackage{multirow, booktabs}
\usepackage{cite}
\usepackage{flushend}

\definecolor{maroon}{cmyk}{0,0.87,0.68,0.32}

\SetKwInput{KwInput}{Input}                
\SetKwInput{KwOutput}{Output}              

\DeclareMathAlphabet{\mathcal}{OMS}{cmsy}{m}{n}

\renewcommand{\CommentSty}[1]{\textnormal{\ttfamily\color{green!50!black}#1}\unskip}

\newcommand*\circled[1]{\tikz[baseline=(char.base)]{
            \node[shape=circle,fill,inner sep=0.6pt] (char) {\textcolor{white}{#1}};}}
            
\title{\LARGE \bf
SePaint: Semantic Map Inpainting via Multinomial Diffusion}
\author{Zheng Chen, Deepak Duggirala, David Crandall, 
Lei Jiang, Lantao Liu 
\thanks{Authors are with Luddy School of Informatics, Computing, and Engineering, Indiana University, Bloomington, IN 47408, USA. \newline
Corresponding author email: {\tt\small zc11@iu.edu}}
}

\begin{document}
\maketitle

\thispagestyle{empty}
\pagestyle{empty}

\begin{abstract}
Prediction beyond partial observations is crucial for robots to navigate in unknown environments because it can provide extra information regarding the surroundings beyond the current sensing range or resolution. In this work, we consider the inpainting of semantic Bird's-Eye-View maps. We propose SePaint, an inpainting model for semantic data based on generative multinomial diffusion. To maintain semantic consistency, we need to condition the prediction for the missing regions on the known regions. We propose a novel and efficient condition strategy, \textit{Look-Back Condition} (LB-Con), which performs one-step look-back operations during the reverse diffusion process. By doing so, we are able to strengthen the harmonization between unknown and known parts, leading to better completion performance. We have conducted extensive experiments on different datasets, showing our proposed model outperforms commonly used interpolation methods in various robotic applications.  
\end{abstract}

\section{Introduction}
\label{sec:intro}
A \textit{dense} semantic map of the surrounding environments can be extremely helpful for autonomous robots to perform tasks  such as 
goal-oriented navigation and autonomous exploration. However, in many cases the robot only has partial observations due to constraints like object occlusions, restricted energy,  limited sensor  resolution or field of view, etc. Incomplete semantic maps containing missing labels can cause significant confusion for downstream planning modules. In this work, we consider the completion problem for Bird's-Eye-View (BEV) semantic maps. 

\begin{figure} [t]
{
\centering
  {\includegraphics[width=\linewidth]{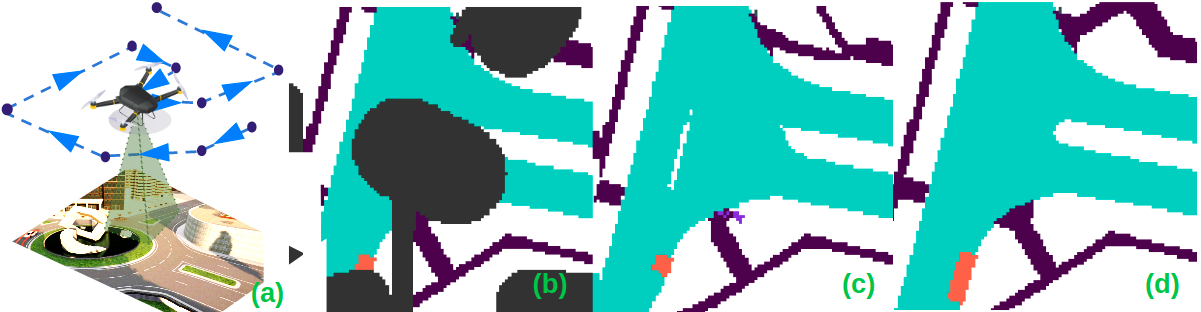}}
\caption{\small (a)  Semantic coverage mapping is performed by a UAV in an urban environment using a downward-facing sensor with a limited field of view. (b) Only a partially revealed map can be obtained if time or energy is tightly constrained (dark color represents  missing regions. See Fig. \ref{fig:nuscenes_qual} for other colors). (c) Our proposed SePaint is able to reason about the underlying semantic structures. (d) Ground truth of the environment.
} 
\label{fig:intro}  
}
\end{figure}

One scenario with partially observed semantic BEV maps is the task of coverage mapping, where the goal of the robot is to map an entire environment. An example of this scenario can be seen in Fig. \ref{fig:intro}-(a), in which an Unmanned Aerial Vehicle (UAV) is assigned to map a city street using a downward-facing sensor. Due to the limited field of view of the  sensor, the UAV can only map a small region at each time step, which may lead to an incomplete map if the overall time or energy budget is constrained.
An example of an incomplete map is in Fig. \ref{fig:intro}-(b), where the dark color represents   regions that are not mapped.

It is also typical that only sparse semantic BEV maps can be obtained when a robot navigates or explores an unknown environment, as a local BEV map is usually constructed for each time step and then accumulated. There are two typical ways to obtain the local BEV semantic map for vision-based perception: (a) Depth-based unprojection and (b) End-to-end prediction. In  depth-based unprojection, one obtains  BEV semantic maps by lifting 2D segmentation images to BEV space using depth information   from either a depth image or a LiDAR scan. In  end-to-end prediction, the BEV map is directly predicted from an RGB image or a video clip by a neural network. These two approaches each have advantages and disadvantages.  End-to-end prediction requires paired RGB images and  ground-truth  BEV maps as training data which can be prohibitively expensive to collect. In contrast,  depth-based unprojection only requires a 2D segmentation dataset to train a segmentation model, as there are currently many more datasets available for 2D segmentation than BEV. Nevertheless, the depth-based unprojection framework can only generate sparse BEV maps due to the limited resolution of depth images or LiDAR scans.

The problem of semantic BEV map completion is similar to  image inpainting and completion \cite{li2022mat, lugmayr2022repaint, wang2022dual}, which aims at filling missing regions within a natural RGB image. The core of both problems is the same --- learning the distribution of semantic maps or  natural RGB images and predicting the absent part based on the known part using the learned distribution. However, an important difference lies in 
the continuity of the data: values in BEV semantic map are label IDs and thus are discrete, while values in natural RGB images represent colors and are continuous. Therefore, we can regard the semantics completion as a discrete image inpainting problem, where semantic labels are sampled from categorical distributions.

In this work, we propose \textbf{SePaint}: a diffusion-based \textbf{Se}mantic in\textbf{Paint}ing framework for categorical data. Our proposed method can be solely based on an unconditionally trained multinomial Denoising Diffusion Probabilistic Model (DDPM). 
To achieve the functionality of inpainting, we condition the inference process on the known parts and propose an efficient condition strategy, \textit{Look-Back Condition} (LB-Con). LB-Con applies noise to both the missing region and the known region, performing a one-step \textit{look-back} operation when reversing. This operation achieves a bidirectional merging of known content and generated content, both in the forward and reverse process. In addition, our method is able to eliminate the requirement of paired input  and corresponding ground truth data. This does not introduce any bias for specific missing patterns and thus allows the model to generalize any masks.  

We validate our proposed framework in different experimental settings with different datasets. First, we show that SePaint is able to inpaint 2D perspective segmentation images with various masks, even though these masks are not seen during training.
Then we apply our approach to semantic maps with different sparsities obtained by a depth-based unprojection. Finally, we demonstrate that our method can also produce multiple samples, and this provides a useful way to quantify the inpainting uncertainty. In all settings, we show our proposed method outperforms  popular interpolation methods such as nearest-neighbor, linear, and cubic interpolation.
In summary, our contributions include:
\begin{itemize}
    \item 
    We  apply a diffusion model to the problem of inpainting for discrete data, and we  show our proposed model has very useful potential in various robotic applications.
    \item
    We propose a novel and efficient condition strategy to harmonize the prediction results with the known regions.
    \item
    We  conduct extensive experiments to validate the advantages of our proposed method over other baselines. We will  make the code of our work public.
\end{itemize}

\section{Related Work}
\label{sec:related}

\textbf{Map Inpainting} aims to predict missing information in a partially observed map, so that  information beyond the robot's perception range can be estimated. This is important to improve the performance and efficiency of a robot's behaviors. There is much existing work attempting to inpaint partial maps. For example, Variational Autoencoders (VAEs) have been used to predict unknown map regions beyond frontiers in a 2D occupancy grid map \cite{shrestha2019learned}. An occupancy anticipation model is proposed in \cite{ramakrishnan2020occupancy} to learn to extrapolate to unseen regions from RGB and depth images. A Conditional Neural Process (CNP) \cite{garnelo2018conditional} is used as the backbone to predict unknown regions based on the context of observed parts \cite{elhafsi2020map}. Occupancy Prediction Network (OPNet) \cite{wang2021learning}, a faster 3D fully-convolutional network  has been proposed to predict 3D occupancy maps, and can be trained using self-supervision. All the above methods belong to data-driven methods and require paired data (observation-gt) to train the map inpainting model. To eliminate the dependency on data, a Low-Rank Matrix Completion (LRMC) based method is proposed to complete a binary map \cite{chen2021efficient}. LRMC is able to achieve both map interpolation and extrapolation on raw poor-quality maps with missing or noisy observations. Although existing work has shown excellent performance, it all targets predicting binary occupancy maps and lacks the capability to predict semantic information.

\textbf{Diffusion Models} 
\cite{sohl2015deep, ho2020denoising} or score-based models \cite{song2019generative, song2020improved, song2020score} are recently emerging powerful alternatives for generative modeling. Essentially, diffusion models are trained to generate an image by denoising  randomly sampled noise in a step-by-step manner. Thanks to principled probabilistic modeling, the objective of diffusion models is surprisingly simple, and the training process is more stable than adversarial training in GANs \cite{NIPS2014_5ca3e9b1}. In addition to the more explicit modeling of the generation process and the more efficient training, diffusion models also show powerful ability on many tasks, such as image translation \cite{su2022dual, rombach2022high}, image editing \cite{meng2021sdedit}, video synthesis \cite{ho2022video}, and language-guided image generation \cite{rombach2022high, kim2022diffusionclip}, and achieve  state-of-the-art  performance on image synthesis \cite{dhariwal2021diffusion} and image inpainting \cite{lugmayr2022repaint}. However, diffusion models suffer from slow inference since they need iterations over many diffusion steps in a chain. Much work has tried to improve the sampling process during inference, including variances during learning \cite{nichol2021improved}, non-Markovian diffusion processes \cite{song2020denoising}, and high-order fast ODE-solvers \cite{lu2022dpm}. Based on prior work, a diffusion-based image inpainting model is proposed in \cite{lugmayr2022repaint}, which  is close to our work and achieves  state-of-the-art  inpainting performance compared with GANs. However, that paper only works on ordinal data such as natural RGB images, thus is not suitable for semantic inpainting.

A framework for discrete diffusion is first described in \cite{song2020denoising}, but no evaluation is conducted. Concurrent to \cite{song2020denoising}, another independent work \cite{hoogeboom2021argmax} also proposes two methods for learning discrete data, Argmax Flows and Multinomial Diffusion, and conducts extensive experiments and shows remarkable generation capability on both segmentation images and natural language. There is other work modeling  discrete data \cite{austin2021structured, campbell2022continuous, dieleman2022continuous, chen2022analog, sun2022score, strudel2022self} but all that work only evaluates their methods on text modeling. In this paper, we use the multinomial diffusion model in \cite{hoogeboom2021argmax} as our backbone for semantic inpainting.

\section{Methodology}
\label{sec:method}
The proposed diffusion-based semantic inpainting framework is detailed in this section. To organize the presentation, we first briefly review  the basic concepts of a standard diffusion model \cite{ho2020denoising}, which usually uses a continuous Gaussian distribution to describe the transition of samples. Then, we describe a discrete extension of the diffusion models, multinomial diffusion \cite{hoogeboom2021argmax}, which can generate categorical data. Finally, we explain the proposed SePaint, a semantic inpainting framework based on the multinomial diffusion model.

\subsection{Continuous Diffusion Models}
\label{sec: continuous}

Given a real data distribution $\mathbf{x}_0 \sim q(\mathbf{x})$, we define a forward diffusion process where a data sample $\mathbf{x}_0$ is transformed in $T$ time steps into a pure-noise image $\mathbf{x}_T$. The joint distribution for all generated newly transformed data can be written as:
\begin{equation}
    \label{eq:1}
    q(\mathbf{x}_{1:T-1}, \mathbf{x}_T|\mathbf{x}_0) = \prod_{t=1}^T q(\mathbf{x}_t | \mathbf{x}_{t-1}),
\end{equation}
where $\mathbf{x}_{1:T-1}$ are intermediate results in the forward process. In each time step $t$, the sample $\mathbf{x}_t$ is computed by adding zero-mean Gaussian noise with variance $\beta_t$ to the sample at the previous time step $\mathbf{x}_{t-1}$ scaled by $\sqrt{1-\beta_t}$. The value of $\beta_t$ is determined by a pre-defined variance schedule. Thus we have the transition distribution as:
\begin{equation}
    \label{eq:2}
    q(\mathbf{x}_t|\mathbf{x}_{t-1}) = \mathcal{N}\left ( \mathbf{x}_t; \sqrt{1-\beta_t}\mathbf{x}_{t-1}, \beta_t \mathbf{I} \right ).
\end{equation}

\begin{figure} 
{
\centering
  {\includegraphics[width=\linewidth]{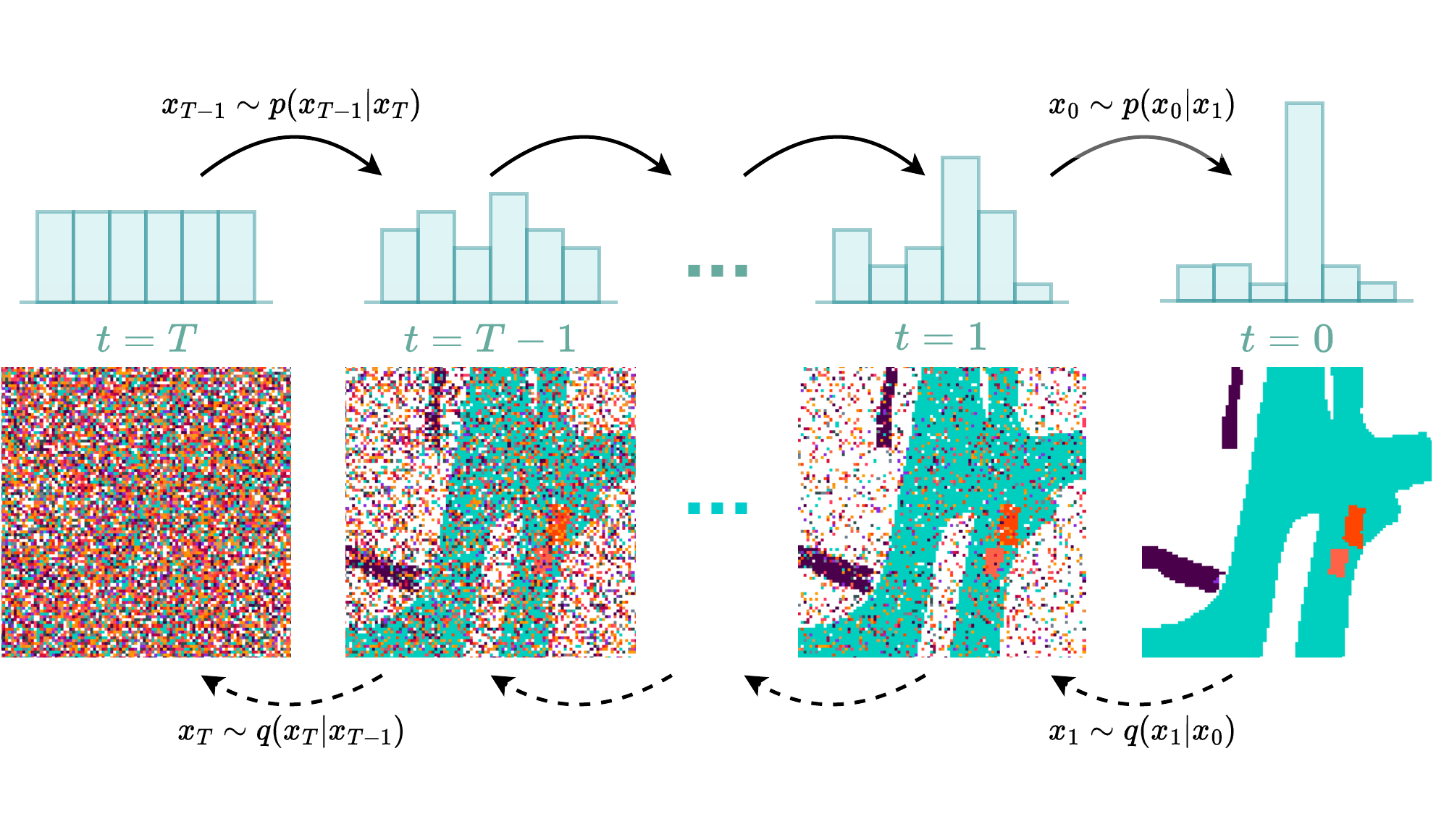}}
\caption{\small Diffusion processes on discrete semantic map data. Top arrows: reverse process. Bottom arrows: forward process.
} 
\label{fig:multi_diff}  
}
\end{figure}

At the final time step $T$, $\mathbf{x}_T$ approaches  an isotropic Gaussian distribution $\mathbf{x}_T \sim \mathcal{N}(0, \mathbf{I})$ when $T\rightarrow\infty$. A nice property of this forward process is that we can obtain sample $\mathbf{x}_t$ at any time step $t$ from $\mathbf{x}_0$ in a closed form:
\begin{equation}
    \label{eq:3}
    q(\mathbf{x}_t|\mathbf{x}_0) = \mathcal{N}(\mathbf{x}_t; \sqrt{\bar{\alpha}}_t \mathbf{x}_0, (1-\bar{\alpha}_t)\mathbf{I}),
\end{equation}
where $\alpha_t = 1 - \beta_t$; $\bar{\alpha}_t = \prod_{i=1}^t \alpha_i$. More details about the derivation from Eq. (\ref{eq:2}) to Eq. (\ref{eq:3}) can be found in \cite{ho2020denoising}.

The core of the diffusion model is to learn a reversed process in Eq. (\ref{eq:2}), i.e., $q(\mathbf{x}_{t-1} | \mathbf{x}_t)$, which is usually approximated by a neural network as follows:
\begin{equation}
    \label{eq:4}
    p_{\theta}(\mathbf{x}_{t-1}|\mathbf{x}_t) = \mathcal{N}(\mathbf{x}_{t-1}; \bm{\mu}_{\theta}(\mathbf{x}_t, t), \bm{\Sigma}_{\theta}(\mathbf{x}_t, t)),
\end{equation}
where $\theta$ denotes the parameters of the neural network that outputs  $\bm{\mu}_{\theta}$ and $\bm{\Sigma}_{\theta}$.

To train the model for estimating Eq. (\ref{eq:4}), we use the variational lower bound to optimize the negative log-likelihood:
\begin{equation}
    \label{eq:5}
    \begin{aligned}
    -\log p_{\theta}(\mathbf{x}_0) &\leq -\log p_{\theta}(\mathbf{x}_0) + \mathcal{D}_{KL}(q(\mathbf{x}_{1:T}|\mathbf{x}_0) || p_{\theta}(\mathbf{x}_{1:T}|\mathbf{x}_0))\\
    &= \mathbb{E}_{q} \left [ \log \frac{q(\mathbf{x}_{1:T}|\mathbf{x}_0)}{p_{\theta}(\mathbf{x}_{0:T})} \right ] \\
    &= L_T + L_{T-1} + \cdots + L_1 + L_0\\
    &= L_{vlb},
    \end{aligned}
\end{equation}
where
\begin{equation}
    \label{eq:6}
    \begin{aligned}
        L_T &= \mathcal{D}_{KL}(q(\mathbf{x}_T|\mathbf{x}_0)||p_{\theta}(\mathbf{x}_T)),\\
        L_{t-1} &= \mathcal{D}_{KL}(q(\mathbf{x}_{t-1}|\mathbf{x}_t, \mathbf{x}_0)||p_{\theta}(\mathbf{x}_{t-1}|\mathbf{x}_t)),\\
        L_0 &= -\log p_{\theta}(\mathbf{x}_0 | \mathbf{x}_1).
    \end{aligned}
\end{equation}
According to \cite{ho2020denoising}, minimizing $L_{vlb}$ is equivalent to minimizing $\sum_{t=2}^{T}L_{t-1}$. Conditioned on $\mathbf{x}_0$, the posterior of $\mathbf{x}_{t-1}$, $q(\mathbf{x}_{t-1}|\mathbf{x}_t, \mathbf{x}_0)$ becomes tractable and has a Gaussian closed form. 
To compute the loss at certain time step $t-1$, we have:
\begin{equation}
    \label{eq:kl}
    L_{t-1} = \mathcal{D}_{KL}(q(\mathbf{x}_{t-1}|\mathbf{x}_t, \mathbf{x}_0) || p_{\theta}(\mathbf{x}_{t-1}|\mathbf{x}_t)).
\end{equation}
With the assumption that the $\bm{\mu}_{\theta}$ has the same form as the mean of $q(\mathbf{x}_{t-1}|\mathbf{x}_t, \mathbf{x}_0)$, the role of the neural network is simplified to estimate the noise $\epsilon_t$ added to $\mathbf{x}_t$ at time step $t$, and the loss term $L_{t-1}$ is parameterized to minimize the difference between the real noise $\epsilon$ and the estimated noise $\epsilon_{\theta}(\mathbf{x}_t, t)$. A simplified version of $L_{t-1}$ is proposed for training:
\begin{equation}
    \label{eq:loss}
    L_{\text{simple}} = \mathbb{E}_{t-1, \mathbf{x}_0, \epsilon} \left [ \left\| \epsilon - \epsilon_{\theta}(\mathbf{x}_{t-1}, t-1) \right\|^2 \right ].
\end{equation}

\subsection{Discrete Diffusion Models}
The multinomial diffusion model \cite{hoogeboom2021argmax} is proposed to generate discrete data, e.g., segmentation images and text, and follows the same development logic as the continuous diffusion model described in Section  \ref{sec: continuous}. The difference lies in the distribution generating the data --- instead of using the continuous Gaussian, the multinomial diffusion model adopts a categorical distribution. In this section, we review key steps in the multinomial diffusion model.

Using the categorical distribution as the transition, the forward diffusion process is:
\begin{equation}
    \label{eq:9}
    q(\mathbf{x}_t|\mathbf{x}_{t-1}) = \mathcal{C}(\mathbf{x}_t; \mathbf{p}=(1-\beta_t)\mathbf{x}_{t-1}+\beta_t / K),
\end{equation}
where $\beta_t$ is the same as Eq. (\ref{eq:2}), and $K$ is the number of classes considered in generation.

Similar to Eq. (\ref{eq:3}), we can use $\mathbf{x}_0$ to predict any intermediate sample $\mathbf{x}_t$:
\begin{equation}
    \label{eq:10}
    q(\mathbf{x}_t|\mathbf{x}_0) = \mathcal{C}(\mathbf{x}_t; \mathbf{p}= \bar{\alpha}_t\mathbf{x}_0 + (1-\bar{\alpha}_t)/K),
\end{equation}
where $\bar{\alpha}_t$ is the same as in Eq. (\ref{eq:3}).

The conditional posterior $q(\mathbf{x}_{t-1}|\mathbf{x}_t, \mathbf{x}_0)$ can be computed in a closed form:
\begin{equation}
    \label{eq:11}
    \begin{aligned}
        q(\mathbf{x}_{t-1}|\mathbf{x}_t, \mathbf{x}_0) = \mathcal{C}(\mathbf{x}_{t-1};\mathbf{p}=\mathbf{p}_{\text{post}}(\mathbf{x}_t, \mathbf{x}_0)),
    \end{aligned}
\end{equation}
where $\mathbf{p}_{\text{post}}(\mathbf{x}_t, \mathbf{x}_0) = \mathbf{p}/\sum_{k=1}^K p_k$ and 
\begin{equation}
    \label{eq:pk}
    \mathbf{p} = \left [ \alpha_t\mathbf{x}_t+(1-\alpha_t)/K \right ] \ \odot \left [ \bar{\alpha}_{t-1}\mathbf{x}_0 + (1 - \bar{\alpha}_{t-1})/K \right ].
\end{equation}

We assume that $p_{\theta}(\mathbf{x}_{t-1}|\mathbf{x}_t)$ has the same form as $q(\mathbf{x}_{t-1}|\mathbf{x}_t, \mathbf{x}_0)$. Thus we have:
\begin{equation}
    \label{eq:13}
    p_{\theta}(\mathbf{x}_{t-1}|\mathbf{x}_t) = \mathcal{C}(\mathbf{x}_{t-1}; \mathbf{p}=\mathbf{p}_{\text{post}}(\mathbf{x}_t, \hat{\mathbf{x}}_0)),
\end{equation}
where $\hat{\mathbf{x}}_0$ is the prediction from the neural network $\hat{\mathbf{x}}_0 = NN_{\theta}(\mathbf{x}_t, t)$. We then are able to obtain the loss for training the multinomial diffusion model:
\begin{equation}
    \label{eq:multi_loss}
    \begin{aligned}
        L_{t-1} &= \mathcal{D}_{KL}(q(\mathbf{x}_{t-1}|\mathbf{x}_t, \mathbf{x}_0) || p_{\theta}(\mathbf{x}_{t-1}|\mathbf{x}_t))\\
        &= \mathcal{D}_{KL}(\mathcal{C}(\mathbf{x}_{t-1}; \mathbf{p}_{\text{post}}(\mathbf{x}_t, \mathbf{x}_0)) || \mathcal{C}(\mathbf{x}_{t-1}; \mathbf{p}_{\text{post}}(\mathbf{x}_t, \hat{\mathbf{x}}_0))).
    \end{aligned}
\end{equation}

An illustration of the multinomial diffusion model can be seen in Fig. \ref{fig:multi_diff}.

\begin{figure}
{
\centering
  {\includegraphics[width=\linewidth]{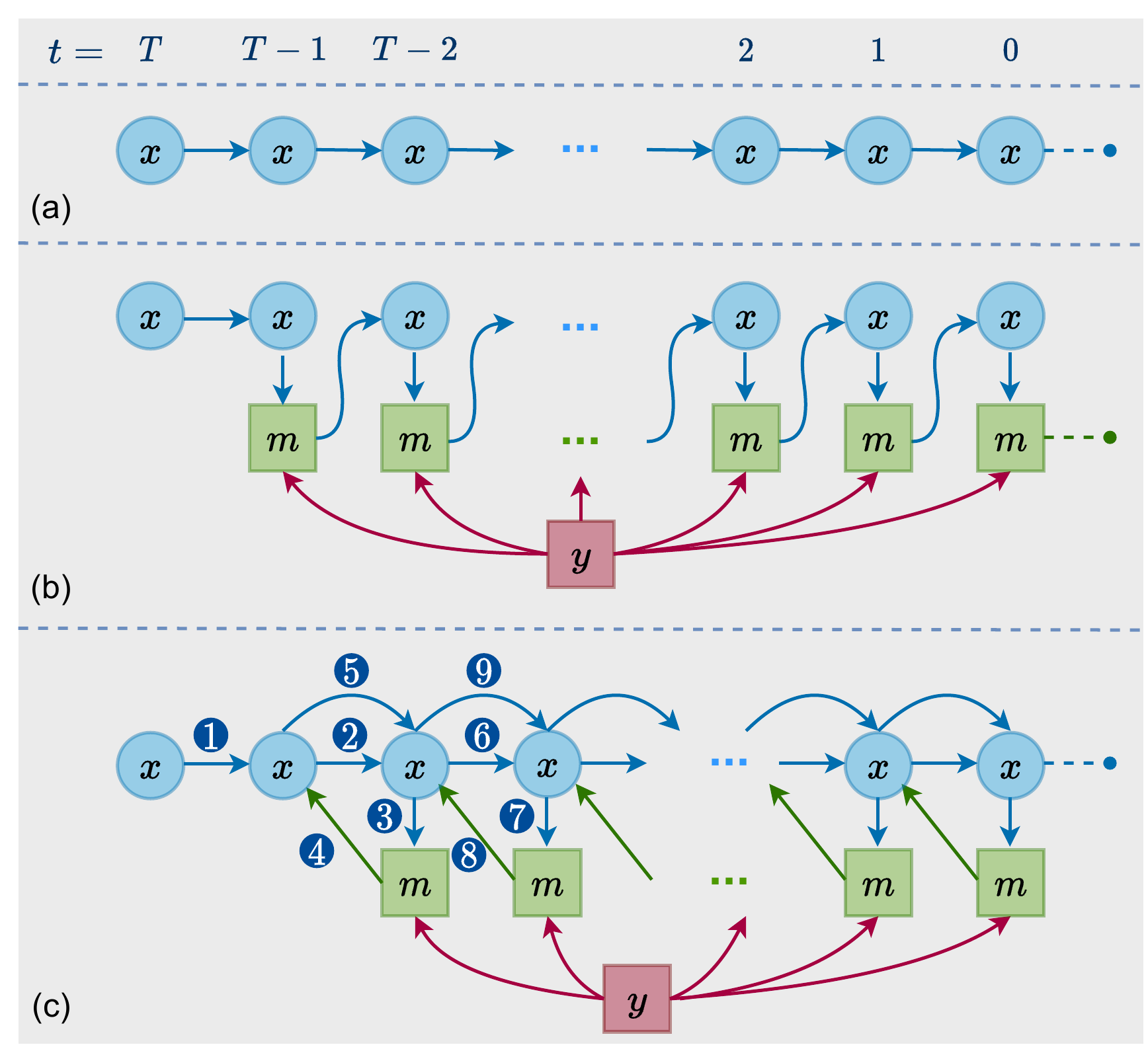}}
\caption{\small Different diffusion-based inference strategies. (a) A standard reverse process. (b) A reverse process with Seq-Con. (c) A reverse process with LB-Con.
} 
\label{fig:sepaint}  
}
\end{figure}

\subsection{Semantic Inpainting}
We denote the results generated during the reverse diffusion process as $\mathbf{x}$, the input semantic map which contains missing regions as $\mathbf{y}$, and the binary mask indicating the known (valued as $1$) and unknown (valued as $0$) as $\mathbf{M}$. The generated content in the missing regions can be extracted by $\mathbf{M}\odot \mathbf{x}$ while the known regions in $\mathbf{y}$ can be obtained by $(\mathbf{1}-\mathbf{M})\odot \mathbf{y}$. We denote the mixture of $\mathbf{x}$ and $\mathbf{y}$ based on the mask $\mathbf{M}$ as $\mathbf{m}$, 
\begin{equation}
    \label{eq:merging_func}
    \begin{aligned}
        \mathbf{m} &= g(\mathbf{x}, \mathbf{y})\\
        &= \mathbf{M}\odot \mathbf{x} + (\mathbf{1}-\mathbf{M})\odot \mathbf{y}.
    \end{aligned}
\end{equation}

Our proposed SePaint depends on an unconditionally pre-trained multinomial diffusion model. The inpainting only occurs during inference such that no mask bias is introduced into the model. Our proposed SePaint is only applied to the inference process, where we treat the trained diffusion model as a generative prior, and is used to predict the data sample at any intermediate time step. In Fig. \ref{fig:sepaint}-(a), we show a standard diffusion-based inference (the reverse process) from a pure noise $\mathbf{x}_T$ to a data sample $\mathbf{x}_0$. In semantic inpainting, to generate missing content, we need to condition the reverse process on the known regions. One straightforward way to achieve this is to convert the known region to a noisy version at each time step and merge the noisy known regions with the generated sample according to the binary mask. We call this simple condition Sequential Condition (\textbf{Seq-Con}), and we show an illustration of an inference process with Seq-Con in Fig. \ref{fig:sepaint}-(b).  Seq-Con can achieve  inpainting and harmonize the content in known regions and unknown regions because the generation of samples $\mathbf{x}_t,~t\in[0, T-2]$ at each time step depends on $\mathbf{m}_{t+1}$, a mixture of content from both regions.
More information on known regions from $\mathbf{y}$ is merged as the inference process goes deeper. However, we notice the Seq-Con is not sufficient to harmonize different regions because the merging only happens in one direction. To address this, we propose a bidirectional merging strategy called Look-Back Condition (\textbf{LB-Con}), an illustration of which can be seen in Fig. \ref{fig:sepaint}-(c). We mark the first $9$ steps of the inference to improve  readability. In contrast to Seq-Con, LB-Con adds noise to the mixture $\mathbf{m}$, performing a one-step forward process, meaning the inference process looks back one step when two reverse steps are made. LB-Con performs  merging in both the forward process and the reverse process, and thus the content in different regions can be better harmonized. Note that sampling from a categorical distribution can be performed using the Gumbel-Max trick \cite{gumbel1954statistical, maddison2014sampling}, which achieves  sampling by adding i.i.d. Gumbel noise to the unnormalized log probabilities and selecting the index with the maximum value. The steps involved in the \texttt{Gumbel-Max} function can be seen in Algorithm 1.

\begin{algorithm}[t]
{\small
  \KwInput{A categorical distribution $\mathcal{C}(\mathbf{p} = \left\{p_i\right\}_{i=1}^K)$; \\~~~~~~~~~$K$ random numbers $\left\{\epsilon_i\right\}_{i=1}^K$ from $U[0, 1]$}
  \KwOutput{A categorical sample $s$}
    $s = \arg\max_i \left ( \log p_i - \log (-\log \epsilon_i) \right )_{i=1}^K$
    
    return $s$
\caption{Gumbel-Max sampling}
} 
\label{algo:gumbel_max}
\end{algorithm}

\begin{algorithm}[t]
{\small
  \KwInput{Partially observed semantic map $\mathbf{y}_0$; Binary mask $\mathbf{M}$}
  \KwOutput{Dense semantic map $\mathbf{x}_0$}
  Import the \texttt{Gumbel-Max} function from Algorithm 1.\\
  $\mathbf{x}_T \leftarrow \texttt{Gumbel-Max}(\mathcal{C}(\mathbf{x}_T; \mathbf{p}=\mathbf{1}/K))$\\
  $\left\{ \epsilon_i \right\}_{i=1}^K, \epsilon_i\sim U[0, 1]$\\
  $\hat{\mathbf{x}}_0 \leftarrow NN_{\theta}(\mathbf{x}_T, T)$\\

  \CommentSty{// Step \circled{1}; Eq. (\ref{eq:sample_x})}
  
  $\mathbf{x}_{T-1} {\small \leftarrow \texttt{Gumbel-Max}(\mathcal{C}(\mathbf{x}_{T-1}; \mathbf{p}=\mathbf{p}_{\text{post}}(\mathbf{x}_T, \hat{\mathbf{x}}_0)), \left\{\epsilon_i\right\}_{i=1}^K))}$
  
  \For{$t= T-2, \cdots, 0$}
    {
        $\left\{ \epsilon_i \right\}_{i=1}^K, \epsilon_i\sim U[0, 1]~if~t>1,~else~ \left\{ \epsilon_i \right\}_{i=1}^K = \mathbf{0}$

        $\hat{\mathbf{x}}_0 \leftarrow NN_{\theta}(\mathbf{x}_{t+1}, t+1)$

        \CommentSty{// Step \circled{2}; Eq. (\ref{eq:sample_x})}
        
        $\mathbf{x}_t \leftarrow  \texttt{Gumbel-Max}(\mathcal{C}(\mathbf{x}_{t}; \mathbf{p}=\mathbf{p}_{\text{post}}(\mathbf{x}_{t+1}, \hat{\mathbf{x}}_0)), \left\{\epsilon_i\right\}_i^K)$

        $\left\{ \epsilon_i \right\}_{i=1}^K, \epsilon_i\sim U[0, 1]~if~t>1,~else~ \left\{ \epsilon_i \right\}_{i=1}^K = \mathbf{0}$

        \CommentSty{// Eq. (\ref{eq:sample_y})}
        
        $\mathbf{y}_t {\small \leftarrow \texttt{Gumbel-Max}(\mathcal{C}(\mathbf{y}_t; \mathbf{p}=\bar{\alpha}_t\mathbf{y}_0+(1-\bar{\alpha}_t)/K), \left\{\epsilon_i\right\}_{i=1}^K)}$

        \CommentSty{// Step \circled{3}; Eq. (\ref{eq:merging_func})}
        
        $\mathbf{m}_t = \mathbf{M}\odot \mathbf{y}_t + (\mathbf{1}-\mathbf{M})\odot \mathbf{x}_t$

        $\left\{ \epsilon_i \right\}_{i=1}^K, \epsilon_i\sim U[0, 1]~if~t>1,~else~ \left\{ \epsilon_i \right\}_{i=1}^K = \mathbf{0}$

        \CommentSty{// Step \circled{4}; Eq. (\ref{eq:look_back})}
        
        $\mathbf{x}_{t+1} {\small \leftarrow \texttt{Gumbel-Max}(\mathcal{C}(\mathbf{x}_{t+1}; \mathbf{p}=(1-\beta_t)\mathbf{m}_t+\beta_t/K), \left\{\epsilon_i\right\}_{i=1}^K)}$

        $\hat{\mathbf{x}}_0 \leftarrow NN_{\theta}(\mathbf{x}_{t+1}, t+1)$

        $\left\{ \epsilon_i \right\}_{i=1}^K, \epsilon_i\sim U[0, 1]~if~t>1,~else~ \left\{ \epsilon_i \right\}_{i=1}^K = \mathbf{0}$

        \CommentSty{// Step \circled{5}; Eq. (\ref{eq:sample_x})}
        
        $\mathbf{x}_t \leftarrow  \texttt{Gumbel-Max}(\mathcal{C}(\mathbf{x}_{t}; \mathbf{p}=\mathbf{p}_{\text{post}}(\mathbf{x}_{t+1}, \hat{\mathbf{x}}_0)), \left\{\epsilon_i\right\}_i^K)$
        
    }
    
    return $\mathbf{x}_0$
\caption{LB-Con for SePaint}
}
\label{algo:sepaint_algo}
\end{algorithm}

We now provide more details for implementing  LB-Con and  present the complete inference pseudo-code for our proposed SePaint. We first describe the merging process at time step $t$, $\mathbf{m}_t = g(\mathbf{x}_t, \mathbf{y}_t)$, where $g$ is the merging function in Eq. (\ref{eq:merging_func}); $\textbf{x}_t$ and $\mathbf{y}_t$ can be obtained based on Eq. (\ref{eq:13}) and Eq. (\ref{eq:10}):
\begin{equation}
    \label{eq:sample_x}
    \begin{aligned}
        \mathbf{x}_t &\sim p_{\theta}(\mathbf{x}_{t}|\mathbf{x}_{t+1})\\
        &\sim \mathcal{C}(\mathbf{x}_{t}; \mathbf{p}=\mathbf{p}_{\text{post}}(\mathbf{x}_{t+1}, \hat{\mathbf{x}}_0)),
    \end{aligned}
\end{equation}
where $\mathbf{x}_0=NN_{\theta}(\mathbf{x}_{t+1}, t+1)$. 

\begin{equation}
    \label{eq:sample_y}
    \begin{aligned}
        \mathbf{y}_t &\sim q(\mathbf{y}_t|\mathbf{y}_0)\\
        &\sim \mathcal{C}(\mathbf{y}_t; \mathbf{p}=\bar{\alpha}_t\mathbf{y}_0+(1-\bar{\alpha}_t)/K),
    \end{aligned}
\end{equation}
where $\mathbf{y}_0$ is the input partial semantic map.

To perform the \textit{look-back} operation in LB-Con, e.g., the step \circled{4} in Fig. \ref{fig:sepaint}-(c), we base the computation on Eq. (\ref{eq:9}) and have:
\begin{equation}
    \label{eq:look_back}
    \begin{aligned}
        \mathbf{x}_{T-1} &\sim q(\mathbf{x}_{T-1}|\mathbf{m}_{T-2})\\
        &\sim \mathcal{C}(\mathbf{x}_{T-1}; \mathbf{p} = (1-\beta_{T-2})\mathbf{m}_{T-2}+\beta_{T-2}/K).
    \end{aligned}
\end{equation}

We use the cosine schedule for $\beta_t$ proposed in \cite{nichol2021improved}. The complete pseudo-code of our proposed LB-Con is shown in Algorithm 2.

\begin{figure} 
{
\centering
  {\includegraphics[width=\linewidth]{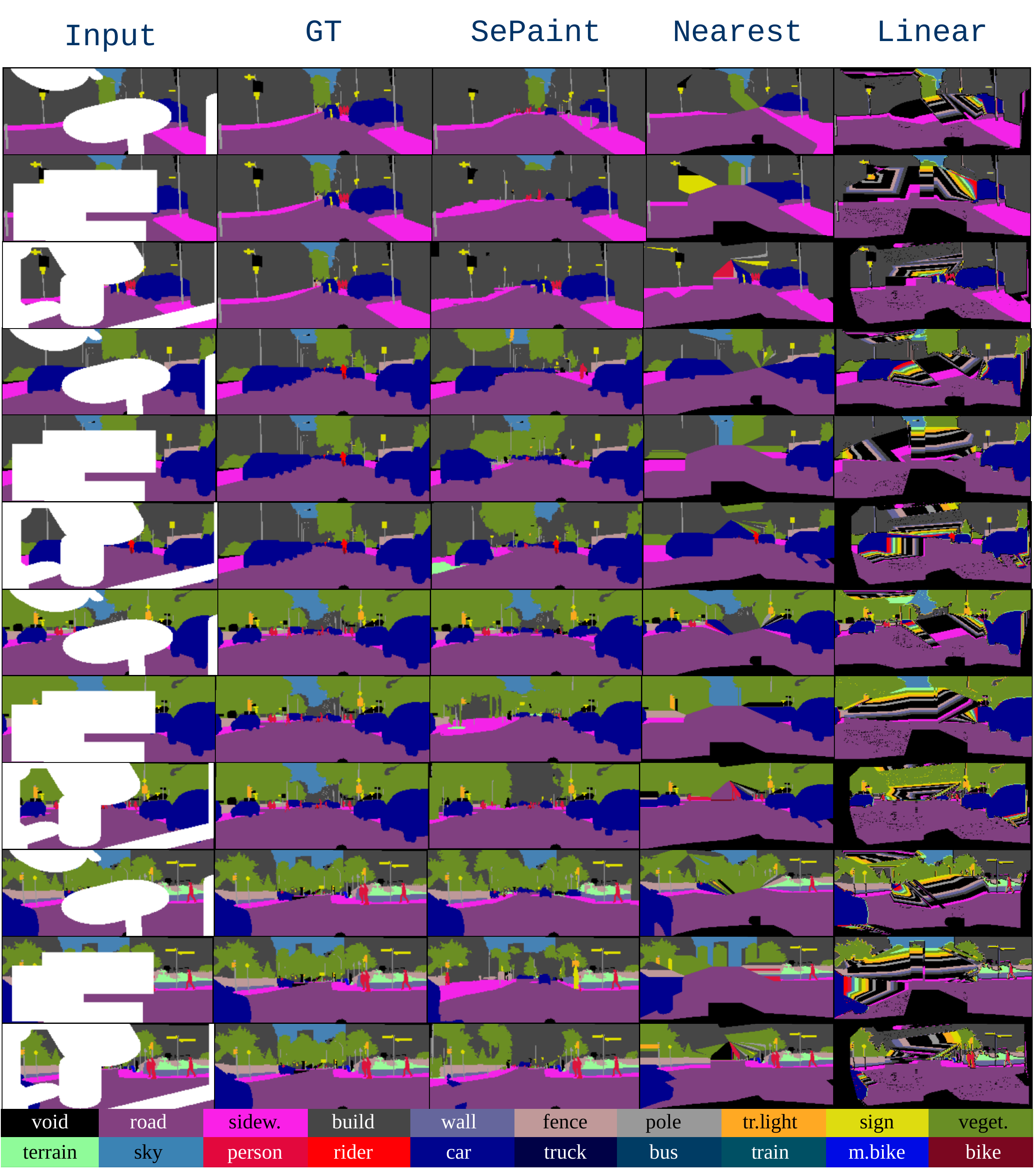}}
\caption{\small Qualitative evaluation on Cityscapes. We use   white to represent the missing regions ($1^{st}$ column).
} 
\label{fig:city_qual}  
}
\end{figure}

\begin{figure} 
{
  \centering
    \subfigure[]
  	{\label{fig:city_miou}\includegraphics[width=0.49\linewidth]{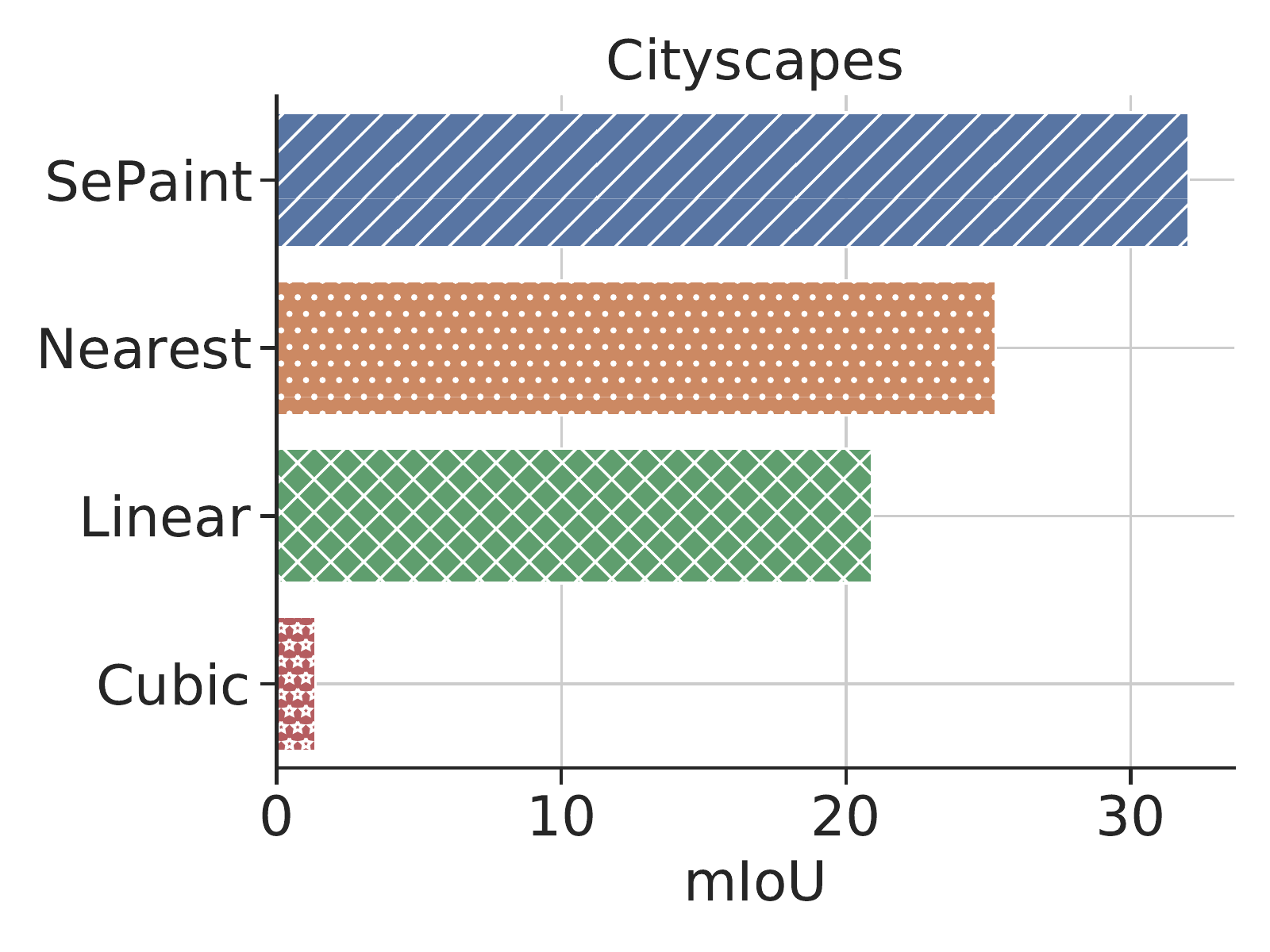}}
  	\subfigure[]
  	{\label{fig:city_acc}\includegraphics[width=0.49\linewidth]{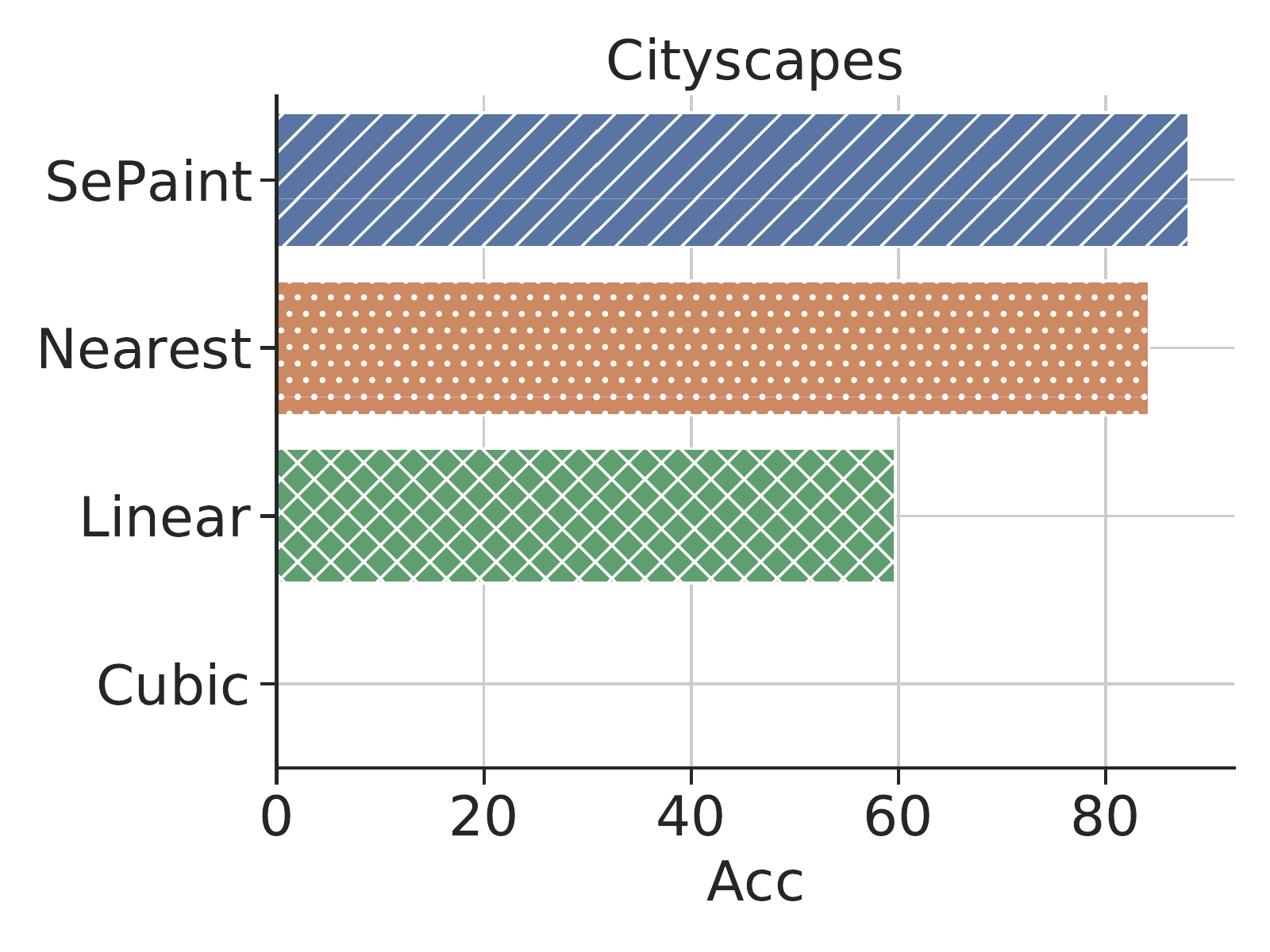}} 
  \caption{\small Quantitative evaluations for different methods on Cityscapes.
  }
\label{fig:city_quan}  
}
\end{figure}

\section{Experiments}
\label{sec:exp}

\subsection{Settings}

We perform extensive experiments on different datasets to demonstrate the advantages of our proposed SePaint. We show that our proposed method can be applied to both perspective data and various forms of BEV data. 

We use two popular datasets for evaluation: Cityscapes \cite{cordts2016cityscapes} and nuScenes \cite{caesar2020nuscenes}.

Cityscapes is a standard 2D semantic segmentation dataset with pixel-wise labels for 19 common categories (e.g., road, sidewalk, tree, person, car, etc.) for images  from different urban environments. To reduce the computational burden, we downsample the images used for training the multinomial diffusion model to $128 \times 256$ using nearest-neighbor interpolation. We use all categories as prediction targets. The official training set ($2,975$ images) is split into $2,500$ images for training and $475$ images for validation. The original testing set is not used. All experimental results reported for Cityscapes are based on the validation images. 

nuScenes is a large-scale dataset for autonomous driving research consisting of high-resolution 3D sensor data collected from a fleet of autonomous vehicles driven in the cities of Boston and Singapore. The dataset contains $1,000$ short video sequences, each frame of which has rich semantic
BEV annotations with 14 classes (e.g., drivable surface, sidewalk, car, truck, bus, etc.). We extract a total of $28,508$ BEV semantic annotations and split those into a training set ($28,008$ images) and a testing set ($500$ images). We use a downsampled size of $100\times100$ during training and testing. 

nuScenes also includes semantic annotations for all sensing points from a 32-beam LiDAR. To approximate unprojected BEV maps from cameras, we project the semantically annotated LiDAR points onto the front camera's image plane, giving us both semantic  and depth images. We use different interpolation methods to densify the projected images: nearest neighbor for semantic images and linear interpolation for depth images. The densification  gives us dense unprojected BEV maps. For our sparse unprojected BEV maps, we directly project the raw annotated LiDAR points to BEV space.



We use the model proposed in \cite{hoogeboom2021argmax} as our unconditional multinomial diffusion model. During training, we use a learning rate of $1\times 10^{-4}$,  batch size of 4 for Cityscapes and 16  for nuScenes, the total number of diffusion steps is set as $4,000$, and we use random horizontal flips as data augmentation. Hand-crafted masks from \cite{lugmayr2022repaint} are used to approximate the missing regions during inference. For baselines we use    nearest-neighbor interpolation, linear interpolation, and cubic interpolation. Two metrics are used to quantitatively evaluate methods: \textbf{Acc}uracy and mean Intersection over Union (m\textbf{IoU}), where  $\mathbf{Acc} = \frac{n_{tp}+n_{tn}}{n_{tp}+n_{tn}+n_{fp}+n_{fn}}$ and $\mathbf{IoU} = \frac{n_{tp}}{n_{tp} + n_{fp} + n_{fn}}$, with $n_{tp}, n_{tn}, n_{fp}$ and $n_{fn}$ as true positives, true negatives, false positives, and false negatives, respectively.

\begin{figure} \vspace{-10pt}
{
\centering
  {\includegraphics[width=\linewidth]{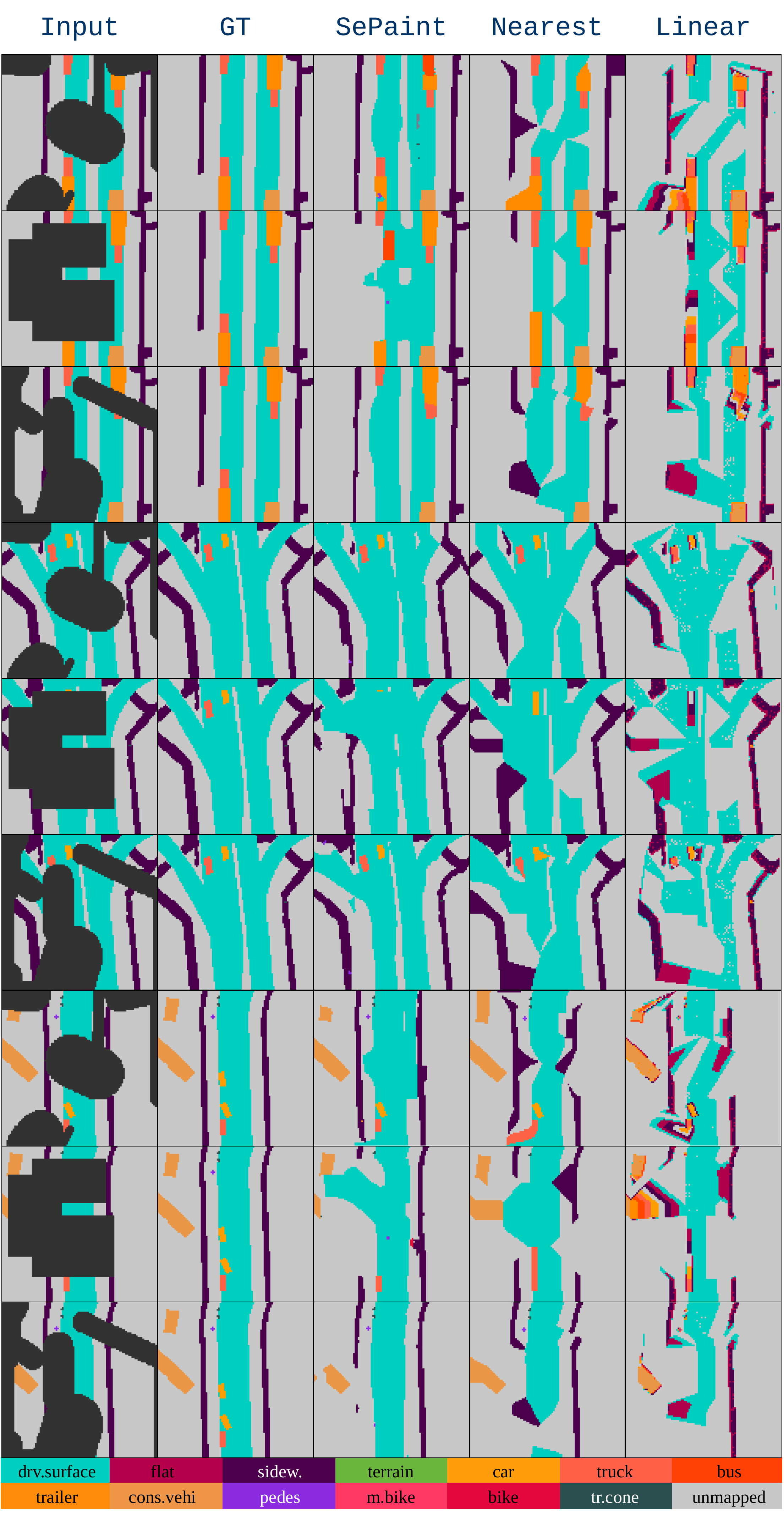}}
\caption{\small Qualitative evaluation on nuScenes. We use   black to represent the missing regions ($1^{st}$ column).
} 
\label{fig:nuscenes_qual}  
}
\end{figure}

\begin{figure} 
{
  \centering
    \subfigure[]
  	{\label{fig:nuscenes_miou}\includegraphics[width=0.49\linewidth]{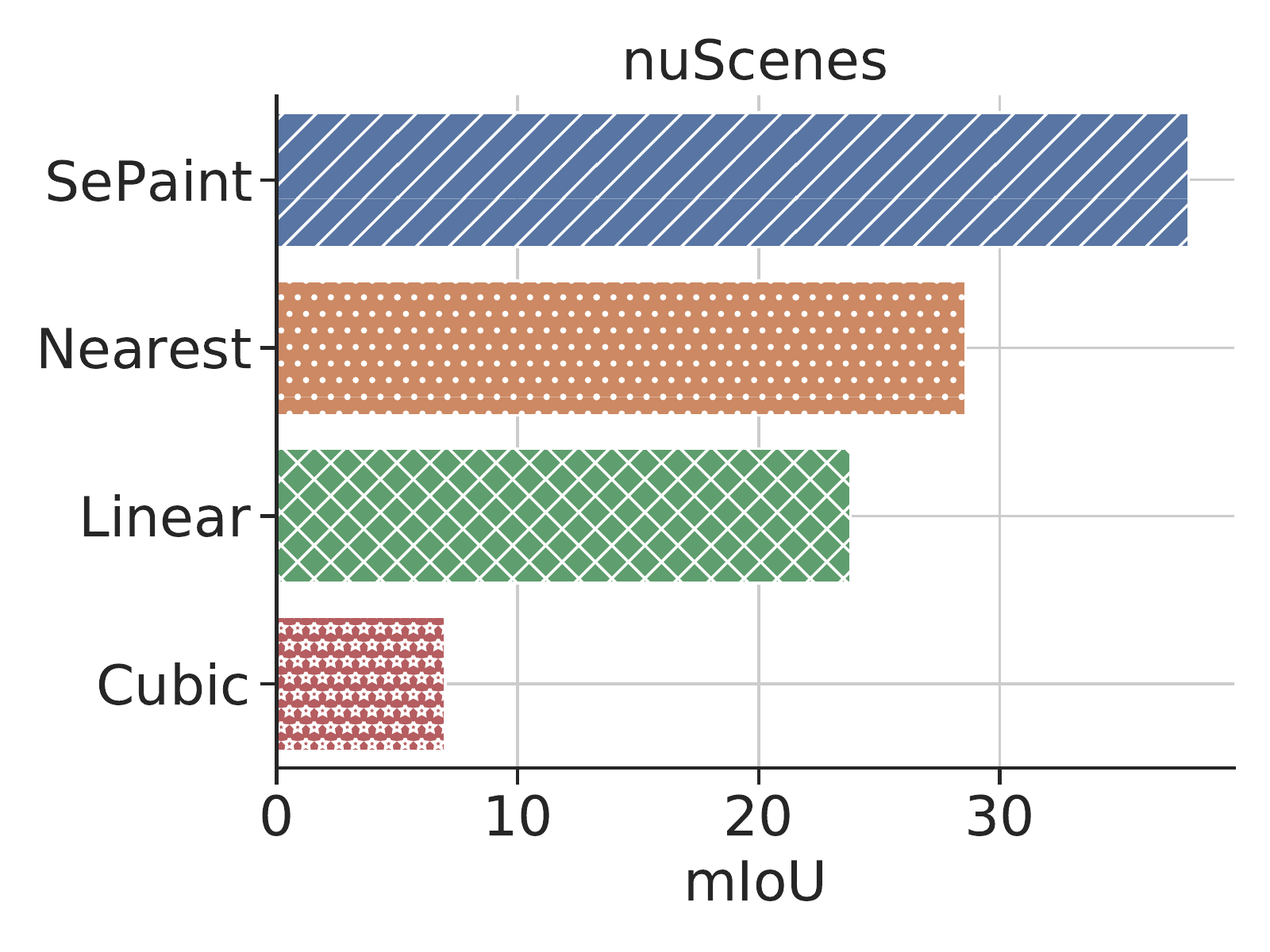}}
  	\subfigure[]
  	{\label{fig:nuscenes_acc}\includegraphics[width=0.49\linewidth]{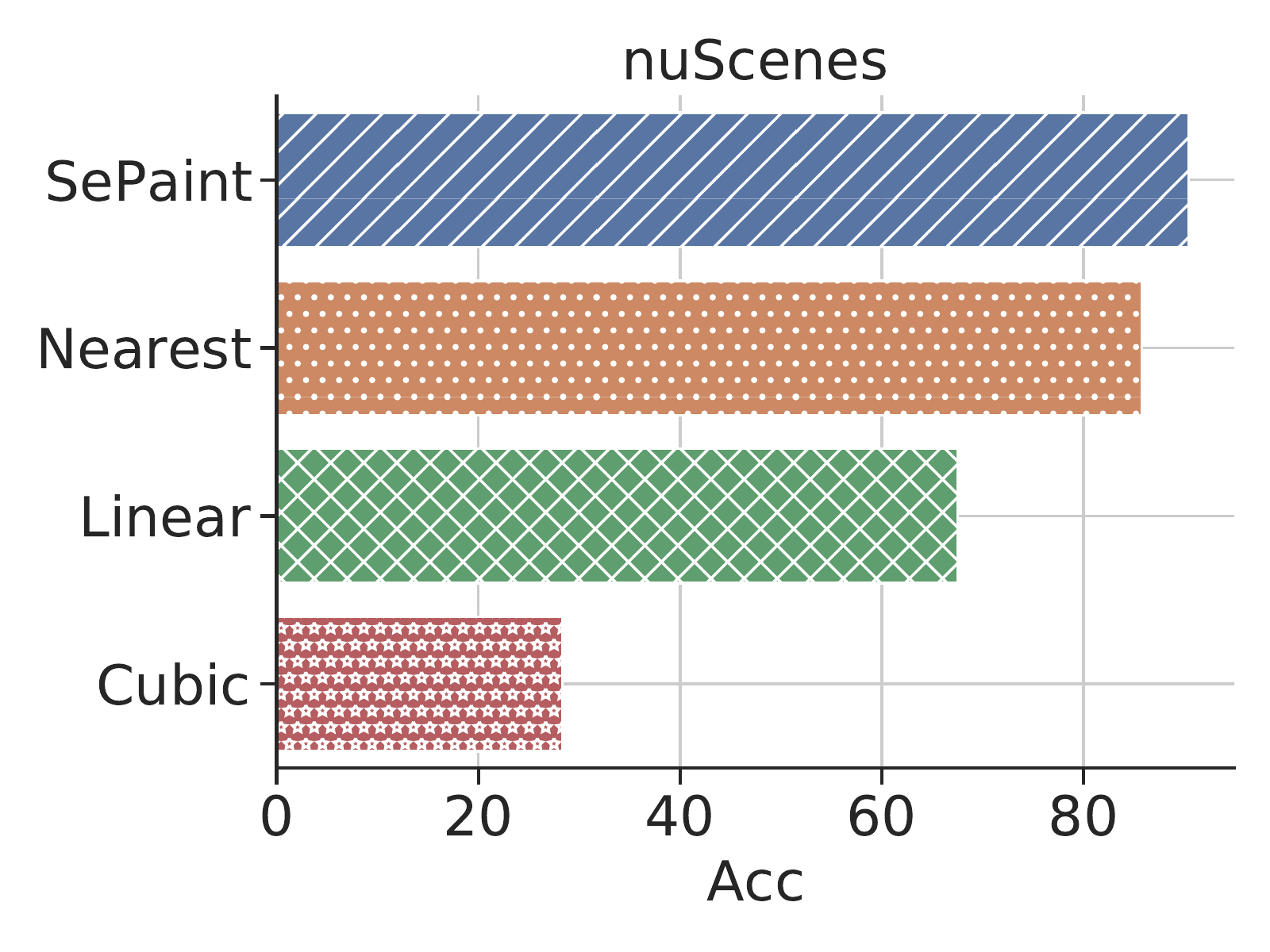}} 
  \caption{\small Quantitative evaluations for different methods on nuScenes.
  }
\label{fig:nuscenes_quan}  
}
\end{figure}

\subsection{Perspective Semantics Completion}
We first validate the effectiveness and advantages of our proposed SePaint on 2D perspective semantic images from Cityscapes. A set of qualitative evaluations can be seen in Fig. \ref{fig:city_qual}. It can be seen that SePaint outperforms all other baselines in terms of prediction accuracy and realistic inference about the scene structure. The reason for this advantage is that SePaint is based on a powerful diffusion generative model and is able to learn the categorical distribution of semantic labels. In contrast, interpolation-based methods only make rigid predictions based on values in known regions and lack prior knowledge of urban scenarios. Among different interpolation-based methods, Nearest-Neighbor has the best performance since semantic labels are discrete. Linear interpolation has the worst performance among all listed methods in Fig. \ref{fig:city_qual}. Cubic interpolation behaves even worse than Linear interpolation (not shown due to space limitations).
Linear/Cubic interpolation can generate new classes out-of-distribution for the current image because there is no causal relation among labels. For example, interpolating between the class $8$ and class $10$ leads to a class of $9$, which is not necessarily the class that is missed and is physically close to the class $8$ and $10$. The true underlying class close to the class $8$ and $10$ can be another possible labels, say $15$. This problem exists because all labels assigned to classes are randomly set --- they are fully independent of each other. Quantitative comparisons can be seen in Fig. \ref{fig:city_quan}, where values of mIoU and Acc for different methods are reported. We run semantic inpainting for all images in the validation set and show the mean of the metric values.

\subsection{BEV Semantics Completion}
In this section, we show comparisons of different methods for completing city semantic BEV maps from nuScenes. We first show qualitative and quantitative results for scenarios of semantic coverage mapping. Then we validate that SePaint is also able to complete unprojected BEV maps where extreme sparsity can happen and strong extrapolations are required.

\begin{figure} 
{
\centering
  {\includegraphics[width=0.85\linewidth]{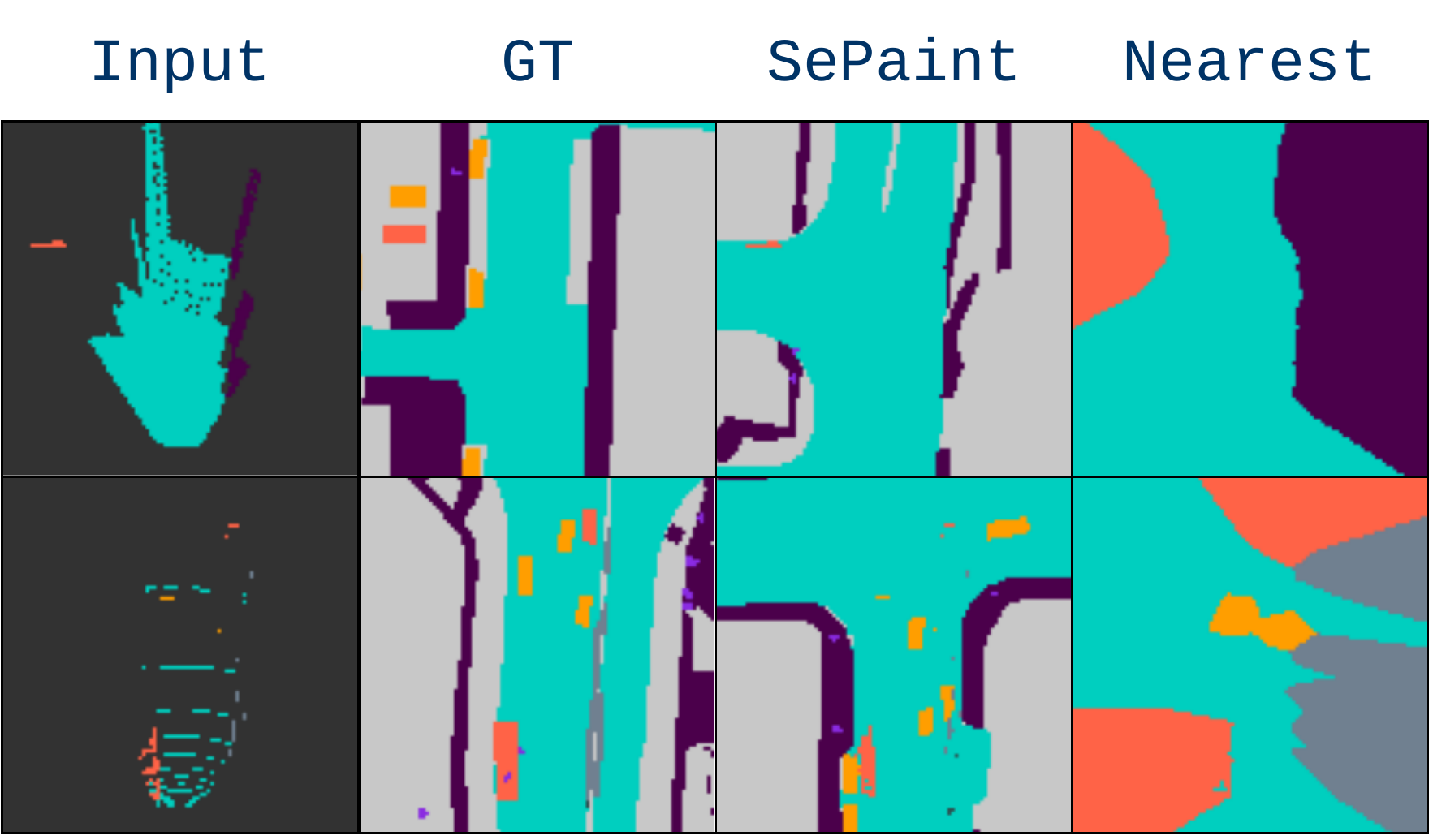}}
\caption{\small Completion for unprojected BEV maps. Unprojected BEV map using Top: depth image; and Bottom: LiDAR points.
} 
\label{fig:raw_map}  
}
\end{figure}

\subsubsection{Completion for Coverage Mapping}
We approximate the partially observed semantic BEV map by applying various binary masks to the semantic BEV data in nuScenes. A set of qualitative evaluations can be seen in Fig. \ref{fig:nuscenes_qual}, where the dark regions in the $1^{st}$ column are missed. Similar to the results for Cityscapes, our proposed method can predict the most semantically-consistent BEV structures and achieve the best performance among all the methods under comparison. Quantitative comparisons can be seen in Fig. \ref{fig:nuscenes_quan}.

\begin{table}
\centering
\caption{Comparison of different methods for completing the unprojected BEV map. } \vspace{-8pt}
\footnotesize
\label{tab:unprojected}
\renewcommand{\arraystretch}{1.1}
\begin{tabular}{cccc}
\hline \hline 
Input map type & Method & mIoU & Acc\\
\hline
\multirow{4}{*}{Dense unprojected map} & Sepaint & \textbf{14.55} & \textbf{53.54}  \\
 & Nearest & 9.77 & 37.16 \\
 & Linear & 10.51 & 42.63 \\
 & Cubic & 7.21 & 42.59 \\
\hline
\multirow{4}{*}{Sparse unprojected map} & Sepaint & \textbf{12.98} & \textbf{59.21} \\
 & Nearest & 7.86 & 21.27 \\
 & Linear & 8.57 & 42.45 \\
 & Cubic & 7.66 & 47.87 \\
\hline \hline
\end{tabular} \vspace{-10pt}
\end{table}

\begin{figure}[t] 
{
\centering
  {\includegraphics[width=\linewidth]{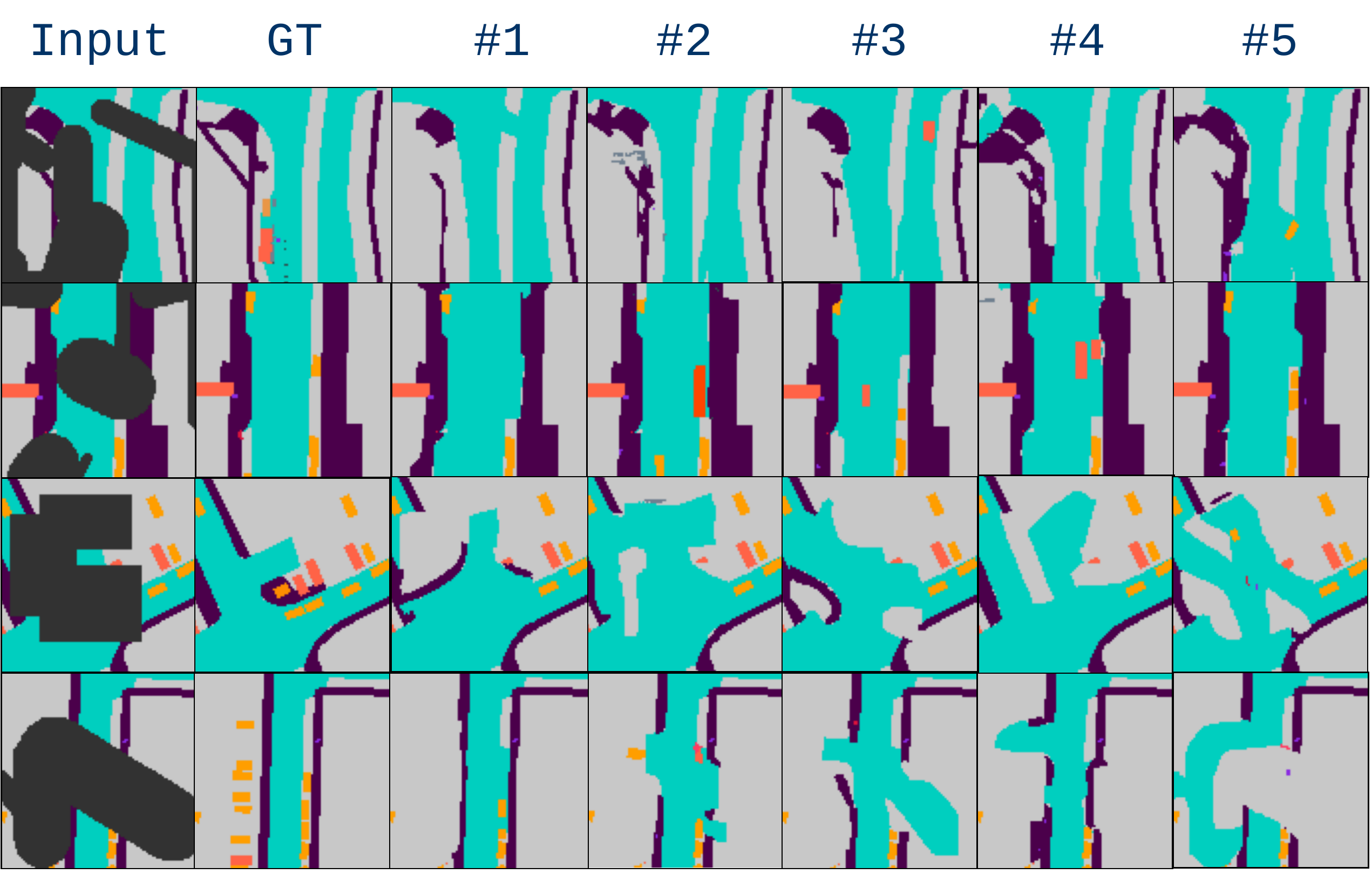}}
\caption{\small Different samples from SePaint.
} 
\label{fig:samples}  
}
\end{figure}

\subsubsection{Completion for Unprojected BEV Map}
One of the commonly used ways for having a BEV semantic map from vision is to unproject a 2D segmentation image using a corresponding depth image or LiDAR points. However, extreme sparsity can be brought down due to the limited resolution of depth images or LiDAR points. We show examples of completing unprojected BEV maps in Fig. \ref{fig:raw_map}. We can see from the $1^{st}$ column that the dark color (missing regions) takes most of the pixels, especially in the LiDAR-based projected map (the bottom row). Even with the presence of extreme sparsity, our method can still \textit{imagine} a highly reasonable structure (see the $3^{rd}$ column of Fig. \ref{fig:raw_map}). In contrast, the representative of interpolation methods for discrete data, Nearest-Neighbor, can only predict missing regions in an overly rigid and simplified manner, leading to unacceptable densified maps. More quantitative details of comparing different methods for the unprojected BEV maps can be found in Table \ref{tab:unprojected}.

\subsection{Multiple Samples}
In this section, we show another unique advantage of our proposed method that can be applied to many different robotic applications such as coverage mapping, goal-oriented navigation, and autonomous exploration: the capability to generate multiple samples, each of which can be consistent and harmonized well with the known regions. We list some examples in Fig. \ref{fig:samples}, where $5$ different samples for each partial map are listed in a row-wise manner. The ability to generate multiple samples can not only provide different possibilities for missing regions but also offer a way to quantify the uncertainty of predictions for the missing parts. The uncertainty can be defined as high if the predictions across different samples are highly varying, and as low if the predictions across different samples are almost the same. The estimated uncertainty can be very useful to robot planning when decision-making on partial maps is needed.

\subsection{Ablation Study}

\begin{table}
\centering
\caption{Ablation study for LB-Con and Seq-Con } \vspace{-8pt}
\footnotesize
\label{tab:ablation}
\renewcommand{\arraystretch}{1.1}
\begin{tabular}{ccccccc}
\hline \hline 
& & City & nuScenes & Dense & Sparse & Samples \\
\hline 
\multirow{2}{*}{mIoU} & LB-Con & \textbf{31.88} & \textbf{37.21} & \textbf{14.55} & \textbf{12.98} & \textbf{35.82} \\
& Seq-Con & 27.65 & 33.11 & 12.34 & 8.56 & 30.98\\
\hline
\multirow{2}{*}{Acc} & LB-Con & \textbf{89.02} & \textbf{92.33} & \textbf{53.54} & \textbf{59.21} & \textbf{89.25} \\
& Seq-Con & 85.88 & 88.39 & 51.14 & 54.27 & 83.66 \\
\hline \hline
\end{tabular}  \vspace{-10pt}
\end{table}

We perform an ablation study to demonstrate the advantages of our proposed LB-Con over the simple Seq-Con. In this study, we evaluate diffusion-based inpaintings with different condition strategies on five experimental settings discussed in the above sections. The settings include the completion for \textbf{City}scapes, \textbf{nuScenes}  (nuScenes), \textbf{Dense} unprojected BEV maps, \textbf{Sparse} unprojected BEV maps, and Generation of multiple \textbf{Samples}. Quantitative evaluations in terms of mIoU and Acc are reported in Table \ref{tab:ablation}, from which we can see our proposed LB-Con has systematic improvements in all tested data compared with the simple Seq-Con.

\section{Conclusion and Future Work}
We present an inpainting model, SePaint, for discrete semantic data based on multinomial diffusion. Our proposed SePaint is a conditional generative model where the generation for missing regions is conditional on the known regions. To harmonize the content of the two regions, we propose a new condition strategy, Look-back Condition (LB-Con), which performs one-step look-back operations during the reverse diffusion process. We have reported extensive experimental results and validated that our proposed model outperforms  commonly used interpolation methods for completing data. To the best of our knowledge, there are no other generative models, such as VAEs or GANs, working on the generation of discrete segmentation images/maps. 

One bottleneck for the current model is the  speed during inference. The current generation is still far from being useful in real-time scenarios, which are commonly required in robotic applications. In the future, we will investigate how to accelerate the discrete multinomial diffusion model and refine our proposed LB-Con such that the inpainting inference speed can be real-time or close to be real-time. 
We will also combine  real robotic data with our model and validate our method in real-world navigation experiments.

\bibliographystyle{unsrt}
\bibliography{ref}

\end{document}